\documentclass{article}

\usepackage[accepted]{icml2025}

\usepackage{microtype}
\usepackage{graphicx}
\usepackage{subfigure}
\usepackage{booktabs}
\usepackage{amsmath,amsfonts,amssymb}
\usepackage{bm}
\usepackage{enumitem}
\usepackage{hyperref}

\usepackage{mathtools}
\usepackage{amsthm}
\usepackage[capitalize,noabbrev]{cleveref}

\theoremstyle{plain}

\theoremstyle{definition}

\theoremstyle{remark}

\usepackage[disable,textsize=tiny]{todonotes}

\begin{document}

\onecolumn  

\makeatletter
\renewcommand{\twocolumn}[1][]{#1}  
\makeatother

\icmltitle{Redefining Robot Generalization Through Interactive Intelligence}

\icmlsetsymbol{equal}{*}

\begin{icmlauthorlist}
\icmlauthor{Sharmita Dey}{}
\end{icmlauthorlist}


\icmlcorrespondingauthor{Sharmita Dey}{contact.deysharmita@gmail.com}

\footnotetext{Corresponding author: Sharmita Dey (\texttt{contact.deysharmita@gmail.com})}

\vskip 0.3in  

\begin{abstract} 

Recent advances in large-scale machine learning have produced high-capacity “foundation models” capable of adapting to a broad array of downstream tasks. While such models hold great promise for robotics, the prevailing paradigm still portrays robots as single, autonomous decision-makers, performing tasks like manipulation and navigation, with limited human involvement. However, a large class of real-world robotic systems, including wearable robotics (e.g., prostheses, orthoses, exoskeletons), teleoperation, and neural interfaces, are semiautonomous, and require ongoing interactive coordination with human partners, challenging single-agent assumptions.  
In this position paper, we argue that robot foundation models must evolve to an \emph{interactive multi-agent} perspective in order to handle the complexities of real-time human-robot co-adaptation. 
We propose a generalizable, neuroscience-inspired architecture encompassing four modules: (1) a multimodal \emph{sensing module} informed by sensorimotor integration principles, (2) an ad-hoc teamwork model reminiscent of joint-action frameworks in cognitive science, (3) a predictive world belief model grounded in internal model theories of motor control, and (4) a memory/feedback mechanism that echoes concepts of Hebbian and reinforcement-based plasticity. Although illustrated through the lens of \emph{cyborg systems}, where wearable devices and human physiology are inseparably intertwined, the proposed framework is broadly applicable to robots operating in semi-autonomous or interactive contexts. By moving beyond single-agent designs, our position emphasizes how foundation models in robotics can achieve a more robust, personalized, and anticipatory level of performance.

\end{abstract}

\section{Introduction}

Over the past few years, the field of AI has been profoundly influenced by \emph{foundation models}, which are large, high-capacity neural networks pre-trained on extensive and heterogeneous datasets \cite{brown2020language, achiam2023gpt}. These models, exemplified by large language models (LLMs) such as GPT-4, and large multimodal models (LMMs) like PaLM-E \cite{driess2023palm}, offer a flexible interface for perception, reasoning, and action. In robotics, foundation models have been applied to unify diverse tasks, e.g., manipulation, navigation, or object recognition, under a single policy \cite{reed2022generalist, brohan2022rt}, often operating in a \emph{single-agent} paradigm where the robot acts autonomously, under minimal human involvement.

However, a large class of real-world robotics, particularly those involving continuous human collaboration, are inherently \emph{multi-agent}. Applications such as teleoperation \cite{lichiardopol2007survey, okamura2004methods, kofman2005teleoperation}, prosthetic devices \cite{best2023data, cimolato2022emg, quintero2017toward, wen2019online, dey2020feasibility, dey2022data}, exoskeletons \cite{molteni2018exoskeleton, rosen2007upper, lo2012exoskeleton, dey2023hard}, neural interfaces \cite{jackson2012neural, donoghue2008bridging, schultz2011neural, vogel2020edan}, brain-computer interfaces \cite{wolpaw2013brain, nicolas2012brain, mcfarland2017eeg, abiri2019comprehensive}, and other semi-autonomous systems \cite{suchan2023assessing, clark2015semi} require ongoing co-adaptation with humans or other participating agents in the environment, rather than isolated, one-shot instructions. In these contexts, the single-agent perspective encounters significant limitations: it struggles to interpret dynamic and evolving user states and fails to handle \emph{non-stationary} factors such as shifting goals, user fatigue, and changing environmental conditions.

Human-interactive robotics and cyborgs, in particular, demand continuous, bidirectional feedback loops between the human user and the device. At each step, the device must integrate physiological signals (e.g., electromyography, joint kinematics), user preferences, and environmental cues to ensure safe and comfortable completion of user commands. Over time, both the human and the device need to learn to function as a coordinated pair. This complexity aligns more closely with \emph{neuroscience-based} perspectives on sensorimotor control, which emphasize dynamic feedback loops, internal predictive models, and adaptive synergy between multiple interacting systems (e.g., brain, muscles, external supports) \cite{wolpert1995internal, kawato1999internal}. Such interactions necessitate a \emph{multi-agent interaction modeling} approach, rather than the traditional model of single-agent autonomy, where both the user and the device function as separate decision-making entities. 

Consequently, this paper takes a position: \emph{future robot foundation models must adopt an interactive multi-agent framework}, explicitly modeling both the robot and its counterpart (human or environment) as actively adapting agents. We illustrate this perspective through wearable robotics or “cyborg systems,” where a user’s physiological signals (e.g., EMG, joint angles) continuously intertwine with the device’s actuation. By incorporating principles from neuroscience, cognitive science, and multi-agent systems, this position paper aims to establish a foundation for next-generation interactive robotics, such as cyborgs and wearable robots, enabling co-adaptation, comfort, and anticipatory control.  We suggest that the widespread adoption of interactive, multi-agent paradigms in robotic foundation models will lead to fundamentally safer, more robust, and more user-centric performance, surpassing what is possible within the single-agent paradigm.

Throughout this paper, we use the term “cyborg systems” (or “wearable robotics”) to refer broadly to human-robot integrated devices, such as prostheses, exoskeletons, brain-computer interfaces, and other assistive technologies that physically couple with the user’s body and neural signals. The paper is organized as follows: Section 2 reviews current robot foundation models, and their single-agent limitations, and introduces multi-agent systems and ad-hoc teamwork. Section 3 explores the requirements of human-interactive scenarios and the shortcomings of single-agent approaches. Section 4 presents our position and proposes an interactive, neuroscience-inspired multi-agent architecture for cyborg systems. Section 5 extends this framework to other human-interactive robotic contexts. Section 7 addresses safety, ethical, and regulatory considerations. The paper concludes in Section 8.

\section{Background}

\subsection{Robot Foundation Models and Their Single-Agent Roots}
The advent of Large Language Models (LLMs) like GPT-4 \cite{achiam2023gpt}, LLaMA \cite{touvron2023llama}, and Vision-Language Models (VLMs) like CLIP \cite{radford2021clip}, BLIP \cite{li2022blip}, BLIP-2 \cite{li2023blip} has significantly advanced robotics by enhancing perception, planning, and action generation capabilities. These models demonstrate exceptional abilities in understanding and generating multimodal data, which are crucial for complex robotic tasks \cite{driess2023palm, brohan2022rt, team2024octo}. By leveraging the robust linguistic capabilities of LLMs, robots can interpret and execute tasks based on natural language commands, eliminating the need for complex programming interfaces. For instance, robots can parse instructions such as "Bring the red cup from the kitchen table" into structured subtasks involving object identification, navigation, and manipulation \cite{brown2020language}.

Despite the impressive capabilities of modern robot foundation models, such as Gato \cite{reed2022generalist} and RT-1 \cite{brohan2022rt}, their predominantly single-agent framework can limit performance in scenarios requiring tight coordination with humans or other agents. These models typically learn policies under the assumption that the robot operates largely on its own, taking in sensory inputs and issuing motor commands without ongoing, interactive feedback from a collaborator or user. Although they have achieved notable results on tasks like manipulation, navigation, and even some language grounding, key shortcomings emerge when real-time collaboration or continuous human guidance is essential. 

\subsubsection{Limitations of Single-Agent Foundation Models in Robotics}

\paragraph{1) Inability to Handle Mid-Task User Corrections.} Gato \cite{reed2022generalist}, for example, was demonstrated on multiple discrete and continuous control tasks, ranging from Atari gameplay to real-world robotic arm manipulation. However, it was not designed to handle situations where a human might intervene mid-task with corrective feedback or dynamically changing instructions (e.g., “Wait, do not place the block there, hand it to me instead.”). As a single-agent learner, Gato follows its end-to-end policy after receiving an initial goal or observation. If the human’s intent shifts \emph{during} task execution, Gato cannot seamlessly incorporate that feedback without externally resetting or retraining the policy. A \emph{multi-agent} perspective, by contrast, would treat the user as a parallel decision-maker; the system would maintain a belief state about the user’s evolving instructions, thus adapting plans in real time rather than requiring full restarts.

\paragraph{2) Lack of Human-Robot “Turn-Taking” in RT-1.} RT-1 \cite{brohan2022rt} demonstrated strong performance across a variety of robot manipulation tasks, incorporating visual inputs and token-based action outputs to execute pick-and-place operations in real-world settings. While it accommodates sensorimotor fusion, the model expects minimal or one-shot user directives, akin to specifying a high-level goal. If a human spontaneously modifies the task (e.g., “Switch from picking an object to now opening a drawer.”), RT-1 has no built-in mechanism to \emph{negotiate} or \emph{re-evaluate} goals on the fly. A multi-agent framework would allow the robot to engage in turn-taking with the human, ask clarifying questions when it detects ambiguous instructions, and update its action policy based on the latest shared understanding.

\paragraph{3) Missed Opportunities for Co-Adaptation.} Both Gato and RT-1 illustrate the single-agent assumption that the robot alone adapts its policy. In scenarios like teleoperation or assistive tasks, however, adaptation is a two-way street: the human also modifies their behavior in response to how the robot is acting, and vice versa. A single-agent viewpoint cannot fully leverage user posture shifts, subtle gestural cues, or real-time user feedback on comfort and safety. By contrast, a multi-agent approach (e.g., ad-hoc teamwork \cite{rahman2021towards, mirsky2022survey}) would explicitly model how the human’s internal state may evolve, whether due to fatigue, changing preferences, or partial completion of sub-goals and modify the robot’s behavior accordingly. This co-adaptive loop can prevent errors (e.g., collisions, user frustration) that arise when the robot rigidly executes a policy absent mutual feedback.

\paragraph{4) Overlooking Collaborative Goal Setting and Preference Tracking.} Another limitation is that single-agent foundation models rarely incorporate long-term \emph{preference tracking} for an external user. While Gato and RT-1 excel at learning general policies from large datasets, they do not maintain a persistent model of a user’s personal constraints or historical preferences (e.g., “User typically prefers lighter grip force on fragile objects” or “User signals discomfort when the end-effector approaches from the left”). In a multi-agent framework, the robot would treat the user’s preferences as a dynamic factor, continuously updating its internal representation as tasks progress and new user feedback arises, thereby improving safety and user satisfaction in real-world contexts \cite{li2021individualized}.

\subsection{Multi-Agent Systems and Ad-Hoc Teamwork}

To address limitations of single-agent paradigms, especially in environments that demand complex, dynamic interactions and collaborative problem-solving, a section of research has advanced towards \emph{multi-agent systems} (MAS) and \emph{ad-hoc teamwork}, frameworks designed to facilitate effective collaboration among multiple entities. This section provides an overview of these paradigms, elucidating their capabilities and advantages over single-agent systems in a general context.

\subsubsection{Multi-Agent Systems}

\emph{Multi-agent systems} consist of multiple interacting agents, each possessing individual goals, capabilities, and decision-making processes \citep{weiss1999multiagent, stone2000multiagent}. These agents can be homogeneous or heterogeneous, cooperative or competitive, and operate within shared or overlapping environments. The primary distinction between MAS and single-agent systems lies in the ability to manage interdependencies and leverage collective intelligence to solve problems that are intractable for individual agents. 

\subsubsection{Ad-Hoc Teamwork}

Building upon the foundation of MAS, \emph{ad-hoc teamwork} \citep{rahman2021towards, barrett2015making, rahman2021towards, melo2016ad} enhances the ability of agents to collaborate effectively without prior coordination, control, or extensive communication. This capability is crucial in environments where agents must form temporary coalitions spontaneously to achieve common objectives, often under conditions of uncertainty and incomplete information \citep{mirsky2022survey}.

\subsubsection{Principles of Ad-Hoc Teamwork}

\begin{itemize} \item \emph{Flexibility and Adaptation:} Agents must adapt to varying team compositions and roles, accommodating new members or the departure of existing ones without significant performance degradation \citep{rahman2021towards}. \item \emph{Implicit Communication:} Effective ad-hoc teamwork can rely on indirect cues and shared environmental information rather than explicit instructions, enabling seamless collaboration with minimal communication overhead \citep{barrett2015making}. \item \emph{Shared Goals and Intentions:} Successful ad-hoc teams align their individual objectives with collective goals, ensuring that all agents work towards a common purpose despite originating from different starting points \citep{barrett2015making}. 
\end{itemize}

\subsubsection{Advantages of Ad-hoc Teamwork Over Single-Agent Systems}

\begin{itemize} \item \emph{Dynamic Adaptation to Human Partners:} In human-robot interactions, ad-hoc teamwork enables robots to adjust their behavior based on real-time human actions and intentions \citep{rahman2021towards}. This adaptability is essential for assistive devices and cyborg systems, where user needs can change rapidly \citep{mirsky2022survey}.

\item \emph{Enhanced Collaboration in Unstructured Environments:} Robots operating in unpredictable settings (e.g., disaster zones or dynamic factory floors) benefit from the ability to form spontaneous coalitions and coordinate responses to new challenges \citep{barrett2017making}.

\item \emph{Improved User Experience:} By enabling more natural and intuitive interactions, often relying on implicit communication, ad-hoc teamwork boosts user satisfaction and trust. Robots are perceived as more reliable and competent partners when they fluidly adapt to human cues.

\item \emph{Adaptive Learning and Co-Evolution:} Agents in multi-agent systems learn from and adapt to one another over time. In human-in-the-loop settings, this means the robot and user can co-evolve, preventing the rigid, static policies typical of single-agent models.

\item \emph{Resilience to Change:} Decentralized or distributed control in multi-agent systems fosters robustness against environmental shifts or new team compositions. When one agent fails or a new user goal emerges, the team can reconfigure itself and maintain overall performance.

\item \emph{Facilitating Human-Robot Synergy:} By explicitly modeling the human’s actions, internal states, and likely next steps, multi-agent systems allow for smoother joint actions, reducing collisions, user discomfort, or misunderstandings about shared goals \citep{barrett2017making}. 
\end{itemize}

By integrating multi-agent and ad-hoc teamwork capabilities, robotic foundation models can achieve a higher level of interaction and collaboration, addressing the fundamental shortcomings of single-agent models. This collaborative framework is essential for developing robots that can engage in real-time, adaptive interactions with humans and other agents, thereby enhancing their functionality and applicability in complex, dynamic environments.

\section{The Human-Device Dyad and Cyborg Systems}

A defining feature of human-interactive robots or cyborg systems (i.e., wearable robotics, including prostheses and exoskeletons) is the \emph{fusion} of human physiology with artificial actuation. In these scenarios, the user (with biological muscles, joints, and neural control) and the robotic device (with actuators, sensors, and algorithms) act as two tightly coupled agents. This bi-directional relationship resonates with sensorimotor loops in neuroscience, where the brain sends motor commands to the musculoskeletal system and receives multisensory feedback to update its internal state \citep{wolpert1998multiple}. 

A single-agent perspective would treat high-level user commands as isolated or sporadic inputs, thereby neglecting the continuous interplay between user and device. This oversight can lead to abrupt control actions, delayed task transitions, or non-personalized responses, especially when the user's biomechanical or cognitive state changes unexpectedly. 
Treating the human and the device as two \emph{agents} with partially observable, evolving states opens the door to ad-hoc collaboration paradigms, wherein:
\begin{itemize}
    \item The device \textit{infers} the user’s intentions and biomechanical constraints from subtle signals like muscle activation patterns.
    \item The user, in turn, adjusts their motor strategies based on feedback from the device’s behavior.
    \item The device maintains and updates an internal model of the user dynamically, for more synergistic control.
\end{itemize}

Even in ostensibly “autonomous” robot applications, there can be hidden interaction partners, such as a human operator providing commands, or an environment whose states change in response to the robot’s actions. Integrating multi-agent interactions into foundation models equips robots with explicit representations of user states, fostering the ability to \emph{predict} and \emph{adapt} continuously based on teammates' models of the world. This integration addresses the fundamental limitations of single-agent, autonomous controllers in human-interactive settings.

\subsection{Alternative to Modern AI-Based Approaches: Finite-State Machines (FSMs)}

Contrasting with modern AI-based controllers, many existing cyborg-like devices utilize \emph{finite-state machines} (FSMs) to determine high-level behaviors such as \textit{standing}, \textit{walking}, or \textit{ascending stairs} based on simple sensor thresholds \citep{tucker2015control}. While FSMs offer straightforward implementation and higher interpretability, they suffer from inherent limitations:

\begin{itemize} \item \emph{Reactive Design}: Transitions typically occur only after sensor thresholds are crossed, leading to potential lag or misclassification when dealing with rapid or subtle user motion changes. \item \emph{Lack of Predictive Modeling}: FSMs generally do not maintain forward-looking estimates of user dynamics or intentions (e.g., anticipating a user’s transition from walking to running before sensor data fully reflects this change), nor do they store long-term user preferences. \item \emph{Minimal Personalization}: Generic thresholds are usually uniform across all users or only slightly tuned, missing opportunities to develop personalized, long-term models of each user’s comfort and biomechanical idiosyncrasies. \end{itemize}

\textbf{Shortfall:} These limitations overlook fundamental insights from neuroscience, such as the role of predictive and adaptive control in biological motor systems. For example, the concept of \emph{efference copy} suggests that the central nervous system forwards internal predictions of movement outcomes to anticipate sensory feedback and modulate subsequent actions \cite{ kawato1999internal}. Similarly, single-agent foundation models lack deeper modeling of user trajectories and internal states, resulting in less fluid and adaptive interactions compared to multi-agent models that incorporate predictive and adaptive control mechanisms.

\subsection{Non-Adaptive User Integration in Single-Agent Models}

Single-agent models often process user inputs as external, episodic commands without maintaining a structured, continuously updated model of the user. This one-directional handling prompts policy decisions without an evolving \emph{teammate model}, which is particularly problematic in wearable robotics where the user’s physiological and behavioral states are dynamic and critically impact device control. Specifically, single-agent, non-adaptive integration falls short in:

\begin{itemize} \item \emph{Tracking User State Trajectories:} Without an explicit internal model of the user's evolving comfort level, fatigue, or motion intent, the device cannot proactively adjust its outputs in response to changes in the user’s gait or stance, potentially leading to overshooting or misalignment of forces. \item \emph{Anticipating Transitions:} If the device treats user commands as static triggers rather than dynamic signals, it may lag behind sudden shifts in user intent, resulting in suboptimal or unsafe responses (e.g., delayed torque adjustments when transitioning to an incline for a prosthesis). \item \emph{Capitalizing on Repeated Interactions:} The absence of memory regarding user-specific preferences or long-term progression hinders the ability to personalize interactions and improve user experience over time. \end{itemize}

In essence, single-agent systems rarely treat the user as a \emph{co-evolving entity} with its own sensorimotor goals. Wearable prosthetics and exoskeletons demand tightly coupled, bidirectional information flow, aligning more closely with a multi-agent perspective that includes explicit or learned models of human states and intentions \cite{rahman2021towards,li2021individualized,mirsky2022survey}. This paradigm shift is especially critical in safety-critical or high-comfort applications, where even minor latency or mismatches in human-device coordination can lead to falls, fatigue, or user frustration.

\section{Position: Embracing an Interactive Multi-Agent Foundational Architecture Inspired by Neuroscience}

We position that \emph{future robot foundation models must adopt an interactive multi-agent framework}, especially for human-in-the-loop or semi-autonomous domains, one that recognizes the user and the robot (e.g., a cyborg prosthetic device) as two interacting agents. We suggest that advanced language or multimodal models should serve not simply as “language-to-action” converters as in current robot foundation models, but as \emph{one component} (a sensing module) within a larger, hierarchical architecture. To concretize this vision, we propose a four-module, neuroscience-inspired approach: 

\begin{itemize}
   \item A \emph{sensing module} that integrates language and multimodal inputs (e.g., EMG, camera data) into structured proposals, mirroring how biological systems merge sensory feedback with high-level goals \citep{pulvermuller2005brain}.
   \item An \emph{ad-hoc teamwork model} \cite{rahman2021towards, li2021individualized, mirsky2022survey} that applies multi-agent collaboration principles, aligning with joint-action and shared intentionality theories in cognitive science \cite{tomasello2003makes, sebanz2006joint}.
    \item A \emph{predictive world belief model} that maintains an internal model of the user and/or collaborating agents' states, enabling anticipatory actions, inspired by motor control theories on forward internal models and predictive coding \cite{wolpert1995internal, wolpert1997computational, wolpert1998multiple, rao1999predictive, millidge2021predictive, denham2020predictive}.
    \item A \emph{memory/feedback subsystem} that stores user-specific preferences and updates policies in a reinforcement-like manner, akin to the role of synaptic plasticity and reinforcement learning in shaping long-term sensorimotor adaptations \cite{dayan2005theoretical}
      
\end{itemize}

\subsection{Module 1: Sensing via Language and Multimodal Inputs}

\paragraph{Neuroscientific Parallels.} Biological organisms integrate sensory cues from diverse modalities (visual, auditory, proprioceptive) to build coherent percepts and guide behavior \cite{stein1993merging}. Similarly, language in humans is thought to interact with high-level planning networks in the brain, providing semantic context and task-relevant goals \cite{pulvermuller2005brain}.

\paragraph{Design Concept.} Our \emph{Sensing Module} leverages language models (e.g., LLaMA \cite{touvron2023llama}) and multimodal models (e.g., PaLM-E \citep{driess2023palm}) to parse: 

\begin{itemize} 

\item \emph{User Commands}: Natural language instructions, which may include explicit directives about speed or comfort, are incorporated into semantically meaningful embeddings. 

\item \emph{Multimodal Inputs}: Visual, auditory, and biomechanical cues (e.g., from onboard cameras, IMUs, or EMG sensors), fused within a multimodal encoder \cite{driess2023palm} or CLIP-like encoders \cite{radford2021clip}. 

\end{itemize}

The Sensing Module outputs a high-level \emph{proposal} about how to adjust the robot control parameters (e.g., torque, stiffness). In essence, it acts like a “multisensory integrator” that also factors in the user’s explicit linguistic preferences. This is reminiscent of how the central nervous system merges sensory feedback with high-level goals to plan motor commands \cite{fuster2002frontal}.

\subsection{Module 2: Ad-hoc Teamwork Model}

\paragraph{Cognitive Science and Joint Action Parallels.} In cognitive science, joint action explores how individuals coordinate tasks by internally representing each other’s goals and actions \cite{sebanz2006joint}. The brain also employs partial models of another person’s internal states in collaborative tasks, often referred to as \emph{shared intentionality} \cite{bratman1992shared}. When humans collaborate, they engage in \emph{shared intentionality}, and mutual understanding and prediction of each other’s intentions, states, and possible actions \cite{bratman1992shared}. Neuroscientific research further reveals that humans utilize a form of theory of mind \cite{baron1994recognition}, allowing one to infer another’s mental states, thereby enabling synchronization in activities such as dancing, carrying a table together, or passing objects. These mechanisms underpin our ability to anticipate partners’ behavior, rapidly adapt to unexpected changes, and maintain coordinated trajectories.

\paragraph{Design Concept.} Robot foundation models can adopt a similar approach, treating the user’s state (fatigue, intention, comfort preference) as an evolving variable to be \emph{modeled}, not just a passive source of commands. This enables collaboration by predicting user behavior in advance, leading to more coordinated and synergistic interactions. Adapting these ideas, our \emph{Ad-hoc Teamwork Model} views the device and user as two agents collaborating under imperfect information. The cumulative function of this module is:

\begin{itemize}
    \item \emph{Intent Inference}: The device (e.g., a leg prosthesis) cannot directly “read” the user’s mind but can \emph{infer} the user’s short-term goals and motor patterns from EMG signals, gait kinematics, and language cues.
    \item \emph{Belief State Maintenance}: Building on the theory of mind \cite{baron1994recognition}, each agent maintains a dynamic belief state about the other. For the prosthesis, this includes anticipating the user’s comfort boundaries (e.g., acceptable torque levels or speed), preferred movement patterns, and likely next moves. By comparing actual outcomes to predictions at each time step, the prosthesis refines its internal model, capturing latent variables such as fatigue onset or changes in user intent.
    
    \item \emph{Refinement of Proposals}: Further, this module refines the high-level proposals from the \emph{sensing module} by combining: 1) \emph{User-centric constraints}, e.g., muscle fatigue or joint stress. 2) \emph{Online collaboration strategies}: e.g., coordinating action output with the user’s shift in stance to achieve stable gait transitions.
\end{itemize}

In multi-agent reinforcement learning (MARL), agents may learn joint policies by sharing partial states \cite{matignon2012independent}. Here, the user’s partial state must be inferred. The device, through repeated interactions, refines its predictive model of the user, aiming for \emph{co-adaptation} rather than a unidirectional control scheme.

\subsection{Module 3: Predictive World Belief Model}

\paragraph{Internal Forward Models in Neuroscience.} A substantial body of research in motor neuroscience underscores the role of \emph{internal forward models}, which predict the sensory outcomes of motor commands, adjusting subsequent behavior in anticipation of future states. \cite{wolpert1998multiple, kawato1999internal}. Predictive coding theories go further, proposing that the brain’s cortex continuously attempts to minimize prediction errors by updating these models \cite{friston2009free}.

\paragraph{Design Concept.} Drawing inspiration from these frameworks, our \emph{Predictive World Belief Model} maintains a probabilistic, forward-looking estimate of both the user’s internal states (muscle fatigue, comfort thresholds, intention shifts) and external conditions (terrain type, slope, potential obstacles). By modeling these dynamic processes over time, this module achieves: 
\begin{itemize} 
\item \emph{Anticipatory control}: The device (e.g., a leg prosthesis) adjusts torque or joint impedance \emph{before} the user’s foot meets a slippery or uneven surface. 
\item \emph{Context-specific transitions}: It can predict that a user’s gait may shift from walking to running or from level-ground to stair ascent, adjusting control parameters accordingly. 
\end{itemize}

Unlike simple threshold-based controllers or purely reactive controllers, a model-based approach predicts the probability distribution of future states. Bayesian strategies, along with predictive coding frameworks \citep{rao1999predictive,friston2009free}, can be leveraged to continuously update such forward-model estimates. 

\subsection{Module 4: Memory \& Feedback for Refinement}

\paragraph{Long-Term Plasticity and Reinforcement in Biology.} Neuroscience literature emphasizes how synaptic plasticity through feedback-driven processes, including dopamine-mediated reinforcement signals, drive \emph{long-term} changes in behavior \cite{gerstner2002spiking, dayan2005theoretical}. For instance, repetitive practice consolidates motor memories that lead to progressive refinement in the brain’s motor representations and improved task efficiency.

\paragraph{Design Concept.} Our \emph{Memory Module} and \emph{Feedback Mechanism} incorporate analogous processes: \begin{itemize} 

\item \emph{Long-term preference storage}: The device (e.g., a leg prosthesis) stores user-specific torque settings, comfort ranges, and frequent command patterns, similar to how repeated exposure to a task solidifies neural pathways in motor learning. 

\item \emph{Reinforcement-based updates}: The device can query the user (“Is this stiffness comfortable?”) or automatically evaluate signals of discomfort, adjusting internal parameters to optimize an objective function (e.g., minimal metabolic cost, subjective comfort). 

\item \emph{Semantic memory bank}: Language-based preferences (“I like a softer ankle when walking on grass”) can be retained as textual or embedding-based knowledge, allowing the device to recall and apply them in the future. \end{itemize}

Hence, the robotic cyborg evolves from a generic device to a personalized “partner,” continuously shaped by the user’s feedback and changing biomechanical needs.

\subsection{Implications for Training and Implementation}

\paragraph{Offline Pre-Training with Diverse Data.}
Much like other foundation models, our system benefits from large-scale pre-training:
\begin{itemize}
    \item \emph{Multimodal corpora:} Recorded EMG, IMU, and camera data across diverse tasks (walking, running, stair-climbing), combined with user commands in natural language.
    \item \emph{Extensive annotation:} Labels for user comfort, stability, and environment factors (terrain type, obstacles) to facilitate robust supervised or self-supervised learning.
\end{itemize}
Challenges include data diversity (users vary widely in gait patterns, limb morphology, or assistance requirements) and the ethical dimensions of collecting sensitive physiological data.

\paragraph{In-Situ Fine-Tuning and Personalization.} 
Aligning with the concept of online plasticity, the system should allow frequent updates based on real-time user feedback to maximize the user’s cumulative comfort or functional metrics (e.g., walking speed, and joint stability). Each device undergoes an online adaptation phase, deploying techniques from continual or reinforcement learning \cite{rolnick2019experience, dayan2005theoretical}. The interacting robotic device thus refines parameters (e.g., stiffness range, torque profiles) while actively interacting with its user in real-world conditions, leveraging the Memory/Feedback mechanism.

\paragraph{Safety and Interpretability.}
Human-interactive robotic devices necessitate robust safety constraints and interpretability.
\begin{itemize}
    \item \emph{Reflex-level safeguards:} The framework should include a “low-level reflex layer,” analogous to spinal reflexes, \cite{wolpaw2002brain} to prevent unsafe actuator outputs.
    
    \item \emph{Explainability tools:} Clinicians and users should be able to understand \textit{why} the device chose specific parameters (e.g., a certain ankle torque), fostering trust and regulatory approval.
\end{itemize}

\section{More Scenarios of Multi-Agent Coordination}

While we emphasize user-device dyads such as cyborg systems, the same multi-agent logic can apply to developing foundation models for extended scenarios such as: \begin{itemize} 

\item \emph{Collaborative Task Environments:} In settings where multiple robotic agents operate alongside human users, such as in industrial manufacturing, disaster response, or healthcare, multi-agent coordination enables seamless collaboration. For instance, in a manufacturing assembly line, multiple robots can dynamically adjust their roles and tasks based on real-time production demands and human worker inputs. Similarly, in disaster response scenarios, a team of drones and ground robots can coordinate their efforts to search for survivors, navigate hazardous terrains, and relay critical information, thereby enhancing the effectiveness and efficiency of the mission \citep{barrett2017making}.
\item \emph{Generalist Models for Smart Homes:} In human-interactive systems, such as personal assistants or smart home environments, multi-agent coordination allows for more personalized interactions. Multiple agents can manage different aspects of the environment (e.g., lighting, climate control, security) while collaboratively adapting to the user's preferences and routines. This can lead to a more cohesive and intuitive user experience, where the system anticipates and responds to the user's needs.

\item \emph{Generalist Synergy Models}: A single user might utilize a lower-limb prosthesis, an upper-limb orthosis, or additional wearable sensors for rehabilitation. Each device can maintain a local instance of the same foundation-model framework, with the potential for \emph{coordinated synergy} if they share a global representation of the user’s state. For instance, an upper-limb exoskeleton might stiffen its elbow joint to assist balance when it detects that the lower-limb prosthesis is transitioning to a challenging slope. Such “whole-body integration” draws on the principle that motor coordination in the human body emerges from distributed neural systems working in tandem \cite{ivry2004neural}. \end{itemize}

\section{Safety, Ethical, and Regulatory Perspectives}

\subsection{Medical Accountability and Clinical Evidence}

Adopting multi-agent foundation models in robotics, particularly within medical and assistive technologies, necessitates stringent compliance with regulatory standards. The integration of multiple autonomous agents introduces complexities in risk assessment and accountability. It is imperative to conduct comprehensive clinical trials and gather substantial evidence to validate the safety and efficacy of these AI-driven control strategies. Ensuring minimal risks of adverse events, such as falls or device malfunctions, and demonstrating reliable performance across diverse user populations is critical for regulatory approval and widespread adoption \citep{bender2021dangers}.

\subsection{Privacy, Data Ownership, and Bias Mitigation}

Multi-agent systems often require extensive collection and processing of physiological and contextual data, increasing privacy concerns. Implementing robust data protection measures, such as on-device processing, federated learning, and anonymized data management, is essential to safeguard user information. Additionally, large-scale models can inadvertently perpetuate biases present in their training data \citep{bender2021dangers}. In healthcare and assistive contexts, such biases could lead to unequal performance across different user groups. Thus, rigorous data curation, bias mitigation strategies, and domain-specific fine-tuning are imperative to ensure equitable and unbiased system performance.

\section{Discussion for Future Research}

\subsection{Toward Adaptive, Personalized Prostheses at Scale}

The proposed methodology heralds a shift from static, command-following prostheses to devices that actively \emph{co-adapt} with their human partners. In practice, widespread adoption relies on: \begin{itemize} \item \emph{Standardized Benchmarks}: Similar to the large-scale robotics benchmarks in manipulation, new testbeds for human-interactive robots, such as, prosthetic synergy are required. Tasks should capture real-world complexity, e.g., irregular outdoor environments, dynamic changes in user fatigue, and prolonged usage scenarios. \item \emph{Open-Source Ecosystems}: Encouraging open data and model sharing can expedite progress. Much like large-scale image or speech datasets have revolutionized computer vision and NLP, similarly sized corpora of wearable robotics data could accelerate foundation-model development. \item \emph{Clinical Partnerships}: Collaboration with clinicians, physical therapists, and user communities can ensure that the system’s objectives align with real-world functional outcomes (e.g., reduced risk of falls, improved metabolic efficiency, subjective comfort). \end{itemize}

\subsection{Active Learning and Human-in-the-Loop Refinement}

Beyond passively receiving user feedback, future prosthetics could proactively query the user to reduce uncertainty: “Would you like increased ankle torque for this incline?” Such \emph{active learning} parallels the process by which humans ask clarifying questions during joint action to refine shared tasks \cite{sebanz2006joint}. In the context of foundation models, these interactions deepen the system’s semantic memory and enhance personalization.

\vspace{-0.3cm}
\section{Conclusion}

We argue that future robot foundation models must be designed from a \emph{multi-agent} perspective to meet the demands of interaction and non-stationarity. Rather than functioning as isolated, single-agent systems, these models should explicitly account for both the robot and its human counterpart (or broader environment) as actively adapting agents. By weaving together insights from neuroscience, cognitive science, and multi-agent systems, next-generation interactive robotics, exemplified by cyborgs and wearable devices, can move beyond one-shot instructions and rigid autonomy. Embracing this multi-agent framework has the potential to ultimately deliver interactive, more robust, and user-centric performance, offering capabilities that surpass the limitations of today’s single-agent paradigms.

\bibliography{example_paper}
\bibliographystyle{icml2025}

\end{document}